\documentclass[11pt,a4paper]{article}

\PassOptionsToPackage{hyphens}{url}\usepackage{hyperref}
\usepackage[hyperref]{acl2020}
\usepackage{times}
\usepackage{latexsym}

\usepackage{microtype}
\usepackage{amssymb}
\usepackage{amsmath}
\usepackage{blindtext}
\usepackage{booktabs}
\usepackage{multirow}
\usepackage{graphicx}
\usepackage{subcaption}
\usepackage{algorithm}
\usepackage{algorithmic}
\usepackage{enumitem}

\aclfinalcopy % Uncomment this line for the final submission
\def\aclpaperid{349} %  Enter the acl Paper ID here

\newcommand\parheader[1]{{\bf \smallskip \noindent #1.}}

\title{Rapidly Bootstrapping a Question Answering Dataset for COVID-19}

\author{Raphael Tang,$^{1}$ Rodrigo Nogueira,$^{1}$ Edwin Zhang,$^{1}$ Nikhil Gupta,$^{1}$ \\ {\bf Phuong Cam,$^{1}$ Kyunghyun Cho,$^{2,3,4,5}$ \and Jimmy Lin}$^1$\\[0.2cm]
$^1$ David R. Cheriton School of Computer Science, University of Waterloo \\
$^2$ Courant Institute of Mathematical Sciences, New York University \\
$^3$ Center for Data Science, New York University \\
$^4$ Facebook AI Research~~
$^5$ CIFAR Associate Fellow \\
}

\date{}

\begin{document}
\maketitle

\begin{abstract}
We present CovidQA, the beginnings of a question answering dataset specifically designed for COVID-19, built by hand from knowledge gathered from Kaggle's \mbox{COVID-19} Open Research Dataset Challenge.
To our knowledge, this is the first publicly available resource of its type, and intended as a stopgap measure for guiding research until more substantial evaluation resources become available.
While this dataset, comprising 124 question--article pairs as of the present version 0.1 release, does not have sufficient examples for supervised machine learning, we believe that it can be helpful for evaluating the zero-shot or transfer capabilities of existing models on topics specifically related to \mbox{COVID-19}.
This paper describes our methodology for constructing the dataset and presents the effectiveness of a number of baselines, including term-based techniques and various transformer-based models.
The dataset is available at \url{http://covidqa.ai/}
\end{abstract}

\section{Introduction}

In conjunction with the release of the COVID-19 Open Research Dataset (CORD-19),\footnote{
\href{https://pages.semanticscholar.org/coronavirus-research}{COVID-19 Open Research Dataset (CORD-19)}
} 
the Allen Institute for AI partnered with Kaggle and other institutions to organize a ``challenge'' around building an AI-powered literature review for COVID-19.\footnote{
\href{https://www.kaggle.com/allen-institute-for-ai/CORD-19-research-challenge/discussion/142298}{COVID-19 Open Research Dataset Challenge}
}
The ``call to arms'' motivates the need for this effort:\ the number of papers related to COVID-19 published per day has grown from around two dozen in February to over 50 by March to over 120 by mid-April.
It is difficult for any human to keep up with this growing literature.

Operationally, the Kaggle effort started with data scientists developing Jupyter notebooks that analyze the literature with respect to a number of predefined tasks (phrased as information needs).
Some of the most promising notebooks were then examined by a team of volunteers---epidemiologists, medical doctors, and medical students, according to Kaggle---who then curated the notebook contents into an up-to-date literature review.
The product\footnote{
\href{https://www.kaggle.com/covid-19-contributions}{COVID-19 Kaggle community contributions}
} 
is organized as a semi-structured answer table in response to questions such as ``What is the incubation period of the virus?'', ``What do we know about viral shedding in urine?'', and ``How does temperature and humidity affect the transmission of 2019-nCoV?''

These answer tables are meant primarily for human consumption, but they provide knowledge for a SQuAD-style question answering dataset, where the input is a question paired with a scientific article, and the system's task is to identify the answer passage within the document.
From the Kaggle literature review, we have manually created CovidQA---which as of the present version 0.1 release comprises 124 question--document pairs.
While this small dataset is not sufficient for supervised training of models, we believe that it is valuable as an in-domain test set for questions related to COVID-19.
Given the paucity of evaluation resources available at present, this modest dataset can serve as a stopgap for guiding ongoing NLP research, at least until larger efforts can be organized to provide more substantial evaluation resources for the community.

\begin{figure*}
{\bf category:} Asymptomatic shedding \\
{\bf subcategory:} Proportion of patients who were asymptomatic \\
{\bf query:} proportion of patients who were asymptomatic \\
{\bf question:} What proportion of patients are asymptomatic? \\[1ex]
{\bf \underline{Answers}} \\[0.5ex]
{\bf id:} \texttt{56zhxd6e} \\
{\bf title:} Epidemiological parameters of coronavirus disease 2019: a pooled analysis of publicly reported individual data of 1155 cases from seven countries \\
{\bf answer:} 49 (14.89\%) were asymptomatic \\[1ex]
{\bf id:} \texttt{rjm1dqk7} \\
{\bf title:} Epidemiological characteristics of 2019 novel coronavirus family clustering in Zhejiang Province
{\bf answer:} 54 asymptomatic infected cases
\caption{Example of a question in CovidQA with two answers.}
\label{example}
\end{figure*}

The contribution of this work is, as far as we are aware, the first publicly available question answering dataset for COVID-19.
With CovidQA, we evaluate a number of approaches for unsupervised (zero-shot) and transfer-based question answering, including term-based techniques and various transformer models.
Experiments show that domain-specific adaptation of transformer models can be effective in an supervised setting, but out-of-domain fine-tuning has limited effectiveness.
Of the models examined, T5~\cite{Raffel:1910.10683:2019} for ranking~\cite{Nogueira_etal_arXiv2020_T5} achieves the highest effectiveness in identifying sentences from documents containing answers.
Furthermore, it appears that, in general, transformer models are more effective when fed well-formed natural language questions, compared to keyword queries.

\section{Approach}
\label{section:approach}

At a high level, our dataset comprises (question, scientific article, exact answer) triples that have been manually created from the literature review page of Kaggle's COVID-19 Open Research Dataset Challenge.
It is easiest to illustrate our approach by example; see Figure~\ref{example}.

The literature review ``products'' are organized into categories and subcategories, informed by ``Tasks'' defined by the Kaggle organizers.\footnote{
\href{https://www.kaggle.com/allen-institute-for-ai/CORD-19-research-challenge/tasks}{COVID-19 Open Research Dataset Challenge Tasks}
}
One such example is ``Asymptomatic shedding'' and ``Proportion of patients who were asymptomatic'', respectively.
The subcategory may or may not be phrased in the form of a natural language question; in this case, it is not.
In the ``question'' of CovidQA, we preserve this categorization and, based on it, manually created both a query comprising of keywords (what a user might type into a search engine) and also a well-formed natural language question.
For both, we attempt to minimize the changes made to the original formulations; see Figure~\ref{example}.

In the Kaggle literature review, for each category/subcategory there is an ``answers table'' that presents evidence relevant to the information need.
Each table is different, and in our running example, the table has columns containing the title of an article that contains an answer, its date, as well as the asymptomatic proportion, age, study design, and sample size.
In this case, according to the site, the entries in the answer table came from notebooks by two data scientists (Ken Miller and David Mezzetti), whose contents were then vetted by two curators (Candler Clawson and Devan Wilkins).

For each row (entry) in the literature review answer table, we began by manually identifying the exact article referenced (in terms of the unique ID) in the COVID-19 Open Research Dataset (\mbox{CORD-19}).
To align with ongoing TREC document retrieval efforts,\footnote{\href{https://ir.nist.gov/covidSubmit/}{TREC-COVID}} we used the version of the corpus from April 10.
Finding the exact document required keyword search, as there are sometimes slight differences between the titles in the answer tables and the titles in CORD-19.
Once we have located the article, we manually identified the exact answer span---a verbatim extract from the document that serves as the answer.
For example, in document \texttt{56zhxd6e} the exact answer was marked as ``49 (14.89\%) were asymptomatic''.

The annotation of the answer span is not a straightforward text substring match in the raw article contents based on the Kaggle answer, but required human judgment in most cases.
For simpler cases in our running example, the Kaggle answer was provided with a different precision.
In more complex cases, the Kaggle answer does not match any text span in the article.
For example, the article may provide the total number of patients and the number of patients who were asymptomatic, but does not explicitly provide a proportion.
In these cases, we used our best judgment---if the absolute numbers were clearly stated in close proximity, we annotated (at least a part of) the exact answer span from which the proportion could be computed.
See Figure~\ref{example} for an example in the case of document \texttt{rjm1dqk7};
here, the total number of patients is stated nearby in the text, from which the proportion can be computed.
In some cases, however, it was not feasible to identify the answer in this manner, and thus we ignored the entry.
Thus, not all rows in the Kaggle answer table translated into a question--answer pair.

A lesson from the QA literature is that there is considerable nuance in defining exact answers and answer contexts, dating back over 20 years~\cite{Voorhees_Tice_TREC8}.
For example, in document \texttt{56zhxd6e}, although we have annotated the answer as ``49 (14.89\%) were asymptomatic'' (Figure~\ref{example}), an argument could be made that ``14.89\%'' is perhaps a better exact span.
We are cognizant of these complexities and sidestep them by using our dataset to evaluate model effectiveness {\it at the sentence level}.
That is, we consider a model correct if it identifies the sentence that contains the answer.
Thus, we only need to ensure that our manually annotated exact answers are (1) proper substrings of the article text, from its raw JSON source provided in CORD-19, and that (2) the substrings do not cross sentence boundaries.
In practice, these two assumptions are workable at an intuitive level, and they allow us to avoid the need to articulate complex annotation guidelines that try to define the ``exactness'' of an answer.

Another challenge that we encountered relates to the scope of some questions.
Drawn directly from the literature review, some questions produced too many possible answer spans within a document and thus required rephrasing.
As an example, for the topic ``decontamination based on physical science'', most sentences in some articles would be marked as relevant.
To address this issue, we deconstructed these broad topics into multiple questions, for example, related to ``UVGI intensity used for inactivating COVID-19'' and ``purity of ethanol to inactivate COVID-19''.

Five of the co-authors participated in this annotation effort, applying the aforementioned approach, with one lead annotator responsible for approving topics and answering technical questions from the other annotators.
Two annotators are undergraduate students majoring in computer science, one is a science alumna, another is a computer science professor, and the lead annotator is a graduate student in computer science---all affiliated with the University of Waterloo.
Overall, the dataset took approximately 23 hours to produce, representing a final tally (for the version 0.1 release) of 124 question--answer pairs, 27 questions (topics), and 85 unique articles.
For each question--answer pair, there are 1.6 annotated answer spans on average.
We emphasize that this dataset, while too small to train supervised models, should still prove useful for evaluating unsupervised or out-of-domain transfer-based models. 

\section{Evaluation Design}
\label{section:design}

The creation of the CovidQA dataset was motivated by a multistage design of end-to-end search engines, as exemplified by our Neural Covidex~\cite{Zhang_etal_arXiv2020_Covidex} for AI2's COVID-19 Open Research Dataset and related clinical trials data.
This architecture, which is quite standard in both academia~\cite{Matveeva_etal_SIGIR2006,Wang_etal_SIGIR2011} and industry~\cite{Pedersen_SIGIR2010,LiuShichen_etal_SIGKDD2017}, begins with keyword-based retrieval to identify a set of relevant candidate documents that are then reranked by machine-learned models to bring relevant documents into higher ranks.

In the final stage of this pipeline, a module would take as input the query and the document (in principle, this could be an abstract, the full text, or some combination of paragraphs from the full text), and identify the most salient passages (for example, sentences), which might be presented as highlights in the search interface~\cite{Lin_etal_INTERACT2003}.

Functionally, such a ``highlighting module'' would not be any different from a span-based question answering system, and thus CovidQA can serve as a test set.
More formally, in our evaluation design, a model is given the question (either the natural language question or the keyword query) and the full text of the ground-truth article in JSON format (from CORD-19).
It then scores each sentence from the full text according to relevance.
For evaluation, a sentence is deemed correct if it contains the exact answer, via substring matching.

From these results, we can compute a battery of metrics.
Treating a model's output as a ranked list, in this paper we evaluate effectiveness in terms of mean reciprocal rank (MRR), precision at rank one (P@1), and recall at rank three (R@3).

\section{Baseline Models}

Let $q:=(q_1, \ldots, q_{L_q})$ be a sequence of query tokens.
We represent an article as $d:=(s_1, \ldots, s_{L_d})$, where $s_i:=(w_1^i, \ldots, w_{L_i}^i)$ is the $i^\text{th}$ sentence in the article.
The goal is to sort $s_1, \ldots, s_{L_d}$ according to their relevance to the query $q$, and to accomplish this, we introduce a scoring function $\rho(q, s_i)$, which can be as simple as BM25 or as complex as a transformer model. 

As there is no sufficiently large dataset available for training a QA system targeted at COVID-19, a conventional supervised approach is infeasible.
We thus resort to unsupervised learning and out-of-domain supervision (i.e., transfer learning) in this paper to evaluate both the effectiveness of these approaches and the usefulness of CovidQA.

\subsection{Unsupervised Methods}

\parheader{Okapi BM25}
For a simple, non-neural baseline, we use the ubiquitous Okapi BM25 scoring function~\cite{Robertson95} as implemented in the Anserini framework~\cite{Yang_etal_SIGIR2017,Yang_etal_JDIQ2018}, with all default parameter settings.
Document frequency statistics are taken from the entire collection for a more accurate estimate of term importance.

\smallskip
\parheader{BERT models}
For unsupervised neural baselines, we considered ``vanilla'' BERT~\cite{devlin-etal-2019-bert} as well as two variants trained on scientific and biomedical articles:\ SciBERT~\cite{Beltagy2019SciBERT} and BioBERT~\cite{lee2020biobert}.
Unless otherwise stated, all transformer models in this paper use the base variant with the cased tokenizer.

For each of these variants, we transformed the query $q$ and sentence $s_i$, separately, into sequences of hidden vectors, $h^q := (h^q_1, \ldots, h^q_{|q|})$ and $h^{s_i}:=(h^{s_i}_{1}, \ldots, h^{s_i}_{|s_i|})$.
These hidden sequences represent the contextualized token embedding vectors of the query and sentence, which we can use to make fine-grained comparisons. 

We score each sentence against the query by cosine similarity, i.e.,
\begin{equation}
\rho(q, s_i) := \max_{j,k} \frac{h^q_j \cdot h^{s_i}_{k}}{\|h^q_j\|\| h^{s_i}_{k}\|}.
\end{equation}
In other words, we measure the relevance of each token in the document by the cosine similarity against all the query tokens, then determine sentence relevance as the maximum contextual similarity at the token level.

\begin{table*}[t]
\centering
%\resizebox{\textwidth}{!}{
\begin{tabular}{ll lll lll}
\toprule[1pt]
\multirow{2}{*}{\#} & \multirow{2}{*}{Model} & \multicolumn{3}{c}{NL Question} & \multicolumn{3}{c}{Keyword Query}  \\
 \cmidrule(lr){3-5}  \cmidrule(lr){6-8}
 & & P@1 & R@3 & MRR & P@1 & R@3 & MRR  \\
\toprule
1 & Random & 0.012 & 0.034 & -- & 0.012 & 0.034 & -- \\
2 & BM25 & 0.150 & 0.216 & 0.243 & 0.150 & 0.216 & 0.243\\
\midrule
3 & BERT (unsupervised) & 0.081 & 0.117 & 0.159 & 0.073 & 0.164 & 0.187\\
4 & SciBERT (unsupervised) & 0.040 & 0.056 & 0.099 & 0.024 & 0.064 & 0.094 \\
5 & BioBERT (unsupervised) & 0.097 & 0.142 & 0.170 & 0.129 & 0.145 & 0.185\\
\midrule 
6 & BERT (fine-tuned on MS MARCO) & 0.194 & 0.315 & 0.329 & \textbf{0.234} & 0.306 & 0.342\\
7 & BioBERT (fine-tuned on SQuAD) & 0.161 & 0.403 & 0.336 & 0.056 & 0.093 & 0.135 \\
8 & BioBERT (fine-tuned on MS MARCO) & 0.194 & 0.313 & 0.312 & 0.185 & 0.330 & 0.322 \\
9 & T5 (fine-tuned on MS MARCO) & \textbf{0.282} & \textbf{0.404} & \textbf{0.415} & 0.210 & \textbf{0.376} & \textbf{0.360} \\
\bottomrule[1pt]
\end{tabular}
%}
\caption{Effectiveness of the models examined in this paper.}
\label{table:results}
\end{table*}

\subsection{Out-of-Domain Supervised Models}

Although at present CovidQA is too small to train (or fine-tune) a neural QA model, there exist potentially usable datasets in other domains.
We considered a number out-of-domain supervised models:

\parheader{BioBERT on SQuAD}
We used BioBERT-base fine-tuned on the SQuAD v1.1 dataset~\cite{rajpurkar2016squad}, provided by the authors of BioBERT.\footnote{\href{https://github.com/dmis-lab/bioasq-biobert}{\texttt{bioasq-biobert} GitHub repo}}
Given the query $q$, for each token in the sentence $s_i$, the model assigns a score pair $(a^i_j, b^i_j)$ for $1 \leq j \leq |s_i|$ denoting the pre-softmax scores (i.e., logits) of the start and end of the answer span, respectively.
The model is fine-tuned on SQuAD to minimize the negative log-likelihood on the correct beginning and ending indices of the answer spans.

To compute relevance, we let $\rho(q, s_i) := \max_{j,k} \max\{a^i_j, b^i_k\}$.
Our preliminary experiments showed much better quality with such a formulation compared to using log-probabilities, hinting that logits are more informative for relevance estimation in span-based models.

\smallskip
\parheader{BERT and T5 on MS MARCO}
We examined pretrained BioBERT, BERT, and T5~\citep{Raffel:1910.10683:2019} fine-tuned on MS MARCO~\citep{nguyen2016ms}.
In addition to BERT and BioBERT to mirror the above conditions, we picked T5 for its state-of-the-art effectiveness on newswire retrieval and competitive effectiveness on MS MARCO~\cite{Nogueira_etal_arXiv2020_T5}.
Unlike the rest of the transformer models, vanilla BERT uses the uncased tokenizer.
To reiterate, we evaluated the base variant for each model here.

T5 is fine-tuned by maximizing the log-probability of generating the output token $\left<\text{true}\right>$ when a pair of query and relevant document is provided while maximizing that of 
the output token $\left<\text{false}\right>$ with a pair of query and non-relevant  document.
See~\citet{Nogueira_etal_arXiv2020_T5} for details.
Once fine-tuned, we use $\log p(\left<\text{true}\right>|q, s_i)$ as the score $\rho(q, s_i)$ for ranking sentence relevance.

To fine-tune BERT and BioBERT, we followed the standard BERT procedure~\cite{devlin-etal-2019-bert} and trained the sequence classification model end-to-end to minimize the negative log-likelihood on the labeled query--document pairs.

\section{Results}

Evaluation results are shown in Table~\ref{table:results}.
All figures represent micro-averages across each question--answer pair due to data imbalance at the question level.
We present results with the well-formed natural language question as input (left) as well as the keyword queries (right).
For P@1 and R@3, we analytically compute the effectiveness of a random baseline, reported in row 1;
as a sanity check, all our techniques outperform it.

The simple BM25 baseline is surprisingly effective (row 2), outperforming the unsupervised neural approaches (rows 3--5) on both natural language questions and keyword queries.
For both types, BM25 leads by a large margin across all metrics.
These results suggest that in a deployed system, we should pick BM25 over the unsupervised neural methods in practice, since it is also much more resource efficient.

Of the unsupervised neural techniques, however, BioBERT achieves the highest effectiveness (row 5), beating both vanilla BERT (row 3) and SciBERT (row 4).
The comparison between these three models allows us to quantify the impact of domain adaptation---noting, of course, that the target domains of both BioBERT and SciBERT may still differ from \mbox{CORD-19}.
We see that BioBERT does indeed improve over vanilla BERT, more so on keyword queries than on natural language questions, with the latter improvement quite substantial (over five points in P@1).
SciBERT, on the other hand, performs worse than vanilla BERT (both on natural language questions and keyword queries), suggesting that its target is likely out of domain with respect to CORD-19.

Our out-of-domain supervised models are much more effective than their unsupervised counterparts, suggesting beneficial transfer effects.
When fine-tuned on MS MARCO, BERT and BioBERT (rows 6 and 8) achieve comparable effectiveness with natural language input, although there is a bit more variation with keyword queries.
This is quite surprising, as BioBERT appears to be more effective than vanilla BERT in the unsupervised setting.
This suggests that fine-tuning on out-of-domain MS MARCO is negating the domain adaptation pretraining in BioBERT.

Comparing BioBERT fine-tuned on SQuAD and MS MARCO (rows 7 vs.\ 8), we find comparable effectiveness on natural language questions; fine-tuning on SQuAD yields lower P@1 but higher R@3 and higher MRR.
On keyword queries, however, the effectiveness of BioBERT fine-tuned on SQuAD is quite low, likely due to the fact that SQuAD comprises only well-formed natural language questions (unlike MS MARCO, which has more diverse queries).

Finally, we observe that T5 achieves the highest overall effectiveness for all but P@1 on keyword queries.
These results are consistent with~\citet{Nogueira_etal_arXiv2020_T5} and provide additional evidence that encoder--decoder transformer models represent a promising new direction for search, question answering, and related tasks.

Looking at the out-of-domain supervised transformer models (including T5) on the whole, we see that models generally perform better with natural language questions than with keyword queries---although vanilla BERT is an outlier here, especially in terms of P@1.
This shows the potential value of users posing well-formed natural language questions, even though they may degrade the effectiveness of term-based matching since well-formed questions sometimes introduce extraneous distractor words, for
 example, ``type'' in a question that begins with ``What type of...'' (since ``type'' isn't usually a stopword).
Thus, in a multistage architecture, the optimal keyword queries used for initial retrieval might differ substantially from the natural language questions fed into downstream neural architectures.
Better understanding of these differences is a potential direction for future research.

\section{Related Work and Discussion}

It is quite clear that CovidQA does not have sufficient examples to train QA models in a supervised manner.
However, we believe that our dataset can be helpful as a test set for guiding NLP research, seeing that there are no comparable resources (as far as we know).
We emphasize that our efforts are primarily meant as a stopgap until the community can build more substantial evaluation resources.

The significant effort (both in terms of money and labor) that is required to create high-quality evaluation products means that their construction constitutes large, resource-intensive efforts---and hence slow.
As a concrete example, for document retrieval, systems for searching CORD-19 were available within a week or so after the initial release of the corpus in mid-March.
However, a formal evaluation effort led by NIST did not kick off until mid-April, and relevance judgments will not be available until early May (more than a month after dozens of systems have been deployed online).
In the meantime, researchers are left without concrete guidance for developing ranking algorithms, unless they undertake the necessary effort themselves to build test collections---but this level of effort is usually beyond the capabilities of individual teams, not to mention the domain expertise required.

There are, of course, previous efforts in building QA datasets in the biomedical domain. The most noteworthy is BioASQ~\cite{tsatsaronis2015overview}, a series of challenges on biomedical semantic indexing and question answering.
BioASQ does provide datasets for biomedical question answering, but based on manual examination, those questions seem quite different from the tasks in the Kaggle data challenge, and thus it is unclear if a more domain-general dataset could be useful for information needs related to COVID-19.
Nevertheless, in parallel, we are exploring how we might rapidly retarget the BioASQ data for our purposes.

There is no doubt that organizations with more resources and access to domain experts will build a larger, higher-quality QA dataset for COVID-19 in the future.
In the meantime, the alternative is a stopgap such as our CovidQA dataset, creating a alternate private test collection, or something like the Mark I Eyeball.\footnote{\href{https://en.wikipedia.org/wiki/Visual_inspection}{Wikipedia: Visual Inspection}}
We hope that our dataset can provide some value to guide ongoing NLP efforts before it is superseded by something better.

There are, nevertheless, a few potential concerns about the current dataset that are worth discussing.
The first obvious observation is that building a QA dataset for COVID-19 requires domain knowledge (e.g., medicine, genomics, etc., depending on the type of question)---yet none of the co-authors have such domain knowledge.
We overcame this by building on knowledge that has already been ostensibly curated by experts with the relevant domain knowledge.
According to Kaggle, notebooks submitted by contributors are vetted by ``epidemiologists, MDs, and medical students'' (Kaggle's own description), and each answer table provides the names of the curators.
A quick check of these curators' profiles does suggest that they possess relevant domain knowledge.
While profiles are self-authored, we don't have any reason to question their veracity.
Given that our own efforts involved mapping already vetted answers to spans within the source articles, we do not think that our lack of domain expertise is especially problematic.

There is, however, a limitation in the current dataset, in that we lack ``no answer'' documents.
That is, all articles are already guaranteed to have the answer in it; the system's task is to find it.
This is an unrealistic assumption in our actual deployment scenario at the end of a multistage architecture~(see Section~\ref{section:design}).
Instead, it would be desirable to evaluate a model's ability to detect when the answer is not present in the document---another insight from the QA literature that dates back nearly two decades~\cite{Voorhees_TREC2001}.
Note this limitation applies to BioASQ as well.

We hope to address this issue in the near future, and have a few ideas for how to gather such ``no answer'' documents.
The two-stage design of the Kaggle curation effort (raw notebooks, which are then vetted by hand) means that results recorded in raw notebooks that do not appear in the final answer tables may serve as a source for such documents.
We have not worked through the details of how this might be operationalized, but this idea seems like a promising route.

\section{Conclusions}

The empirical nature of modern NLP research depends critically on evaluation resources that can guide progress.
For rapidly emerging domains, such as the ongoing COVID-19 pandemic, it is likely that no appropriate domain-specific resources are available at the outset.
Thus, approaches to rapidly build evaluation products are important.
In this paper, we present a case study that exploits fortuitously available human-curated knowledge that can be manually converted into a resource to support automatic evaluation of computational models.
Although this process is still rather labor intensive, it would be valuable in the future to generalize our efforts into a reproducible methodology for rapidly building information access evaluation resources, so that the community can respond to the next crisis in a timely manner.

\section{Acknowledgments}

This work would not have been possible without the efforts of all the data scientists and curators who have participated in Kaggle's COVID-19 Open Research Dataset Challenge.
Our research is supported in part by the Canada First Research Excellence Fund, the Natural Sciences and Engineering Research Council (NSERC) of Canada, CIFAR AI and COVID-19 Catalyst Grants, NVIDIA, and eBay.
We'd like to thank Kyle Lo from AI2 for helpful discussions and Colin Raffel from Google for his assistance with T5.

\bibliographystyle{acl_natbib}
\bibliography{main}

\begin{thebibliography}{19}
\expandafter\ifx\csname natexlab\endcsname\relax\def\natexlab#1{#1}\fi

\bibitem[{Bajaj et~al.(2016)Bajaj, Campos, Craswell, Deng, Gao, Liu, Majumder,
  McNamara, Mitra, Nguyen, Rosenberg, Song, Stoica, Tiwary, and
  Wang}]{nguyen2016ms}
Payal Bajaj, Daniel Campos, Nick Craswell, Li~Deng, Jianfeng Gao, Xiaodong Liu,
  Rangan Majumder, Andrew McNamara, Bhaskar Mitra, Tri Nguyen, Mir Rosenberg,
  Xia Song, Alina Stoica, Saurabh Tiwary, and Tong Wang. 2016.
\newblock {MS} {MARCO}: {A} human generated {MAchine} {Reading} {COmprehension}
  dataset.
\newblock \emph{arXiv:1611.09268}.

\bibitem[{Beltagy et~al.(2019)Beltagy, Lo, and Cohan}]{Beltagy2019SciBERT}
Iz~Beltagy, Kyle Lo, and Arman Cohan. 2019.
\newblock {S}ci{BERT}: A pretrained language model for scientific text.
\newblock In \emph{Proceedings of the 2019 Conference on Empirical Methods in
  Natural Language Processing and the 9th International Joint Conference on
  Natural Language Processing (EMNLP-IJCNLP)}, pages 3615--3620, Hong Kong,
  China.

\bibitem[{Devlin et~al.(2019)Devlin, Chang, Lee, and
  Toutanova}]{devlin-etal-2019-bert}
Jacob Devlin, Ming-Wei Chang, Kenton Lee, and Kristina Toutanova. 2019.
\newblock {BERT}: Pre-training of deep bidirectional transformers for language
  understanding.
\newblock In \emph{Proceedings of the 2019 Conference of the North {A}merican
  Chapter of the Association for Computational Linguistics: Human Language
  Technologies, Volume 1 (Long and Short Papers)}, pages 4171--4186,
  Minneapolis, Minnesota.

\bibitem[{Lee et~al.(2020)Lee, Yoon, Kim, Kim, Kim, So, and
  Kang}]{lee2020biobert}
Jinhyuk Lee, Wonjin Yoon, Sungdong Kim, Donghyeon Kim, Sunkyu Kim, Chan~Ho So,
  and Jaewoo Kang. 2020.
\newblock {BioBERT}: a pre-trained biomedical language representation model for
  biomedical text mining.
\newblock \emph{Bioinformatics}, 36(4):1234--1240.

\bibitem[{Lin et~al.(2003)Lin, Quan, Sinha, Bakshi, Huynh, Katz, and
  Karger}]{Lin_etal_INTERACT2003}
Jimmy Lin, Dennis Quan, Vineet Sinha, Karun Bakshi, David Huynh, Boris Katz,
  and David~R. Karger. 2003.
\newblock What makes a good answer? {The} role of context in question
  answering.
\newblock In \emph{Proceedings of the Ninth IFIP TC13 International Conference
  on Human-Computer Interaction (INTERACT 2003)}, pages 25--32, {Z\"{u}rich,}
  Switzerland.

\bibitem[{Liu et~al.(2017)Liu, Xiao, Ou, and Si}]{LiuShichen_etal_SIGKDD2017}
Shichen Liu, Fei Xiao, Wenwu Ou, and Luo Si. 2017.
\newblock Cascade ranking for operational e-commerce search.
\newblock In \emph{Proceedings of the 23rd ACM SIGKDD International Conference
  on Knowledge Discovery and Data Mining (SIGKDD 2017)}, pages 1557--1565,
  Halifax, Nova Scotia, Canada.

\bibitem[{Matveeva et~al.(2006)Matveeva, Burges, Burkard, Laucius, and
  Wong}]{Matveeva_etal_SIGIR2006}
Irina Matveeva, Chris Burges, Timo Burkard, Andy Laucius, and Leon Wong. 2006.
\newblock High accuracy retrieval with multiple nested ranker.
\newblock In \emph{Proceedings of the 29th Annual International ACM SIGIR
  Conference on Research and Development in Information Retrieval (SIGIR
  2006)}, pages 437--444, Seattle, Washington.

\bibitem[{Nogueira et~al.(2020)Nogueira, Jiang, and
  Lin}]{Nogueira_etal_arXiv2020_T5}
Rodrigo Nogueira, Zhiying Jiang, and Jimmy Lin. 2020.
\newblock Document ranking with a pretrained sequence-to-sequence model.
\newblock \emph{arXiv:2003.06713}.

\bibitem[{Pedersen(2010)}]{Pedersen_SIGIR2010}
Jan Pedersen. 2010.
\newblock Query understanding at {Bing}.
\newblock In \emph{Industry Track Keynote at the 33rd Annual International ACM
  SIGIR Conference on Research and Development in Information Retrieval (SIGIR
  2010)}, Geneva, Switzerland.

\bibitem[{Raffel et~al.(2019)Raffel, Shazeer, Roberts, Lee, Narang, Matena,
  Zhou, Li, and Liu}]{Raffel:1910.10683:2019}
Colin Raffel, Noam Shazeer, Adam Roberts, Katherine Lee, Sharan Narang, Michael
  Matena, Yanqi Zhou, Wei Li, and Peter~J. Liu. 2019.
\newblock Exploring the limits of transfer learning with a unified text-to-text
  transformer.
\newblock In \emph{arXiv:1910.10683}.

\bibitem[{Rajpurkar et~al.(2016)Rajpurkar, Zhang, Lopyrev, and
  Liang}]{rajpurkar2016squad}
Pranav Rajpurkar, Jian Zhang, Konstantin Lopyrev, and Percy Liang. 2016.
\newblock {SQ}u{AD}: 100,000+ questions for machine comprehension of text.
\newblock In \emph{Proceedings of the 2016 Conference on Empirical Methods in
  Natural Language Processing}, pages 2383--2392, Austin, Texas.

\bibitem[{Robertson et~al.(1995)Robertson, Walker, Hancock-Beaulieu, Gatford,
  and Payne}]{Robertson95}
Stephen~E. Robertson, Steve Walker, Micheline Hancock-Beaulieu, Mike Gatford,
  and A.~Payne. 1995.
\newblock Okapi at {TREC-4}.
\newblock In \emph{Proceedings of the Fourth Text REtrieval Conference
  (TREC-4)}, pages 73--96, Gaithersburg, Maryland.

\bibitem[{Tsatsaronis et~al.(2015)Tsatsaronis, Balikas, Malakasiotis, Partalas,
  Zschunke, Alvers, Weissenborn, Krithara, Petridis, Polychronopoulos,
  Almirantis, Pavlopoulos, Baskiotis, Gallinari, {Arti\'{e}res}, Ngomo, Heino,
  Gaussier, Barrio-Alvers, Schroeder, Androutsopoulos, and
  Paliouras}]{tsatsaronis2015overview}
George Tsatsaronis, Georgios Balikas, Prodromos Malakasiotis, Ioannis Partalas,
  Matthias Zschunke, Michael~R. Alvers, Dirk Weissenborn, Anastasia Krithara,
  Sergios Petridis, Dimitris Polychronopoulos, Yannis Almirantis, John
  Pavlopoulos, Nicolas Baskiotis, Patrick Gallinari, Thierry {Arti\'{e}res},
  Axel-Cyrille~Ngonga Ngomo, Norman Heino, Eric Gaussier, Liliana
  Barrio-Alvers, Michael Schroeder, Ion Androutsopoulos, and Georgios
  Paliouras. 2015.
\newblock An overview of the {BioASQ} large-scale biomedical semantic indexing
  and question answering competition.
\newblock \emph{BMC Bioinformatics}, 16(1):138.

\bibitem[{Voorhees(2001)}]{Voorhees_TREC2001}
Ellen~M. Voorhees. 2001.
\newblock Overview of the {TREC} 2001 question answering track.
\newblock In \emph{Proceedings of the Tenth Text REtrieval Conference (TREC
  2001)}, pages 42--51, Gaithersburg, Maryland.

\bibitem[{Voorhees and Tice(1999)}]{Voorhees_Tice_TREC8}
Ellen~M. Voorhees and Dawn~M. Tice. 1999.
\newblock The {TREC}-8 question answering track evaluation.
\newblock In \emph{Proceedings of the Eighth Text REtrieval Conference
  (TREC-8)}, pages 83--106, Gaithersburg, Maryland.

\bibitem[{Wang et~al.(2011)Wang, Lin, and Metzler}]{Wang_etal_SIGIR2011}
Lidan Wang, Jimmy Lin, and Donald Metzler. 2011.
\newblock A cascade ranking model for efficient ranked retrieval.
\newblock In \emph{Proceedings of the 34th Annual International ACM SIGIR
  Conference on Research and Development in Information Retrieval (SIGIR
  2011)}, pages 105--114, Beijing, China.

\bibitem[{Yang et~al.(2017)Yang, Fang, and Lin}]{Yang_etal_SIGIR2017}
Peilin Yang, Hui Fang, and Jimmy Lin. 2017.
\newblock {Anserini}:\ enabling the use of {Lucene} for information retrieval
  research.
\newblock In \emph{Proceedings of the 40th Annual International ACM SIGIR
  Conference on Research and Development in Information Retrieval (SIGIR
  2017)}, pages 1253--1256, Tokyo, Japan.

\bibitem[{Yang et~al.(2018)Yang, Fang, and Lin}]{Yang_etal_JDIQ2018}
Peilin Yang, Hui Fang, and Jimmy Lin. 2018.
\newblock {Anserini}:\ reproducible ranking baselines using {Lucene}.
\newblock \emph{Journal of Data and Information Quality}, 10(4):Article 16.

\bibitem[{Zhang et~al.(2020)Zhang, Gupta, Nogueira, Cho, and
  Lin}]{Zhang_etal_arXiv2020_Covidex}
Edwin Zhang, Nikhil Gupta, Rodrigo Nogueira, Kyunghyun Cho, and Jimmy Lin.
  2020.
\newblock Rapidly deploying a neural search engine for the {COVID-19} {Open}
  {Research} {Dataset}: Preliminary thoughts and lessons learned.
\newblock \emph{arXiv:2004.05125}.

\end{thebibliography}

\end{document}